\documentclass[conference]{IEEEtran}
\IEEEoverridecommandlockouts
\usepackage{cite}
\usepackage{hyperref}
\usepackage{amsmath,amssymb,amsfonts}
\usepackage{algorithmic}
\usepackage{graphicx}
\usepackage{textcomp}
\usepackage{xcolor}
\usepackage[table]{xcolor}
\def\BibTeX{{\rm B\kern-.05em{\sc i\kern-.025em b}\kern-.08em
    T\kern-.1667em\lower.7ex\hbox{E}\kern-.125emX}}
\begin{document}

\title{Interpretable Physics-Informed Load Forecasting for U.S. Grid Resilience: SHAP-Guided Ensemble Validation in Hybrid Deep Learning Under Extreme Weather\\
% {\footnotesize \textsuperscript{*}Note: Sub-titles are not captured in Xplore and
% should not be used}
% \thanks{Identify applicable funding agency here. If none, delete this.}
}

% \author{
% \IEEEauthorblockN{1\textsuperscript{st} Sajib Debnath}
% \IEEEauthorblockA{\textit{O\&M Analytics, AES Clean Energy} \\
% \textit{The AES Corporation}\\
% Louisville, CO 80027, USA \\
% sajib.debnath@aes.com
% }
% \and

% \IEEEauthorblockN{2\textsuperscript{nd} Md Abubakkar}
% \IEEEauthorblockA{\textit{dept. of Computer Sciences} \\
% \textit{Midwestern State University}\\
% Dallas, TX, USA \\
% mabubakkar@ieee.org}
% \and
% \IEEEauthorblockN{3\textsuperscript{rd} Md. Uzzal Mia}
% \IEEEauthorblockA{\textit{dept. of Information and Communication Engineering} \\
% \textit{Pabna University of Science and Technology}\\
% Rajapur, Pabna, Bangladesh \\
% uzzal.220605@s.pust.ac.bd}
% }

\author{
\IEEEauthorblockN{
Md Abubakkar\textsuperscript{1},
Sajib Debnath\textsuperscript{2},  
Md. Uzzal Mia\textsuperscript{3}
}
\IEEEauthorblockA{
\textsuperscript{1}Dept. of Computer Science, Midwestern State University, Dallas, TX, USA \\
\textsuperscript{2}O\&M Analytics, AES Clean Energy, The AES Corporation, Louisville, CO, USA \\
\textsuperscript{3}Dept. of Information and Communication Engineering, Pabna University of Science and Technology, Bangladesh \\
Email: mabubakkar@ieee.org, sajib.debnath@aes.com, uzzal.220605@s.pust.ac.bd
}
}

\maketitle

\begin{abstract}
Accurate short-term electricity load forecasting is a cornerstone of U.S. grid reliability; however, prevailing deep learning models remain opaque, limiting operator trust during extreme weather. A unified, interpretable, physics-informed ensemble framework is proposed, integrating a Convolutional Neural Network (CNN) branch for local feature extraction and a Transformer branch for long-range dependency modeling; the branches are fused through a validation-optimized weighted ensemble and regularized by a physics-informed loss derived from the piecewise parabolic temperature-demand relationship of the Electric Reliability Council of Texas (ERCOT) system. Post-hoc interpretability is provided through SHapley Additive exPlanations (SHAP) with the DeepExplainer backend, yielding global and event-level attributions. Using eight years of ERCOT hourly load data (2018-2025) fused with Automated Surface Observing System (ASOS) records from three Texas stations, the framework achieves 713 MW MAE, 812 MW RMSE, and 1.18\% MAPE on the test window. For Hampel-flagged extreme events, MAPE falls by 20.7\% relative to its Transformer branch and by 40.5\% relative to its CNN branch; an ablation confirms that the parabolic and ramp constraints drive a 14.7\% RMSE reduction. SHAP analysis reveals a regime shift: temperature dominates under normal operation, whereas wind speed and precipitation become more influential during cold fronts and heatwaves.
\end{abstract}

\begin{IEEEkeywords}
Load forecasting, Explainable AI (XAI), SHAP, Physics-informed neural networks, Ensemble learning, ERCOT, Extreme weather, Smart grid resilience, CNN, Transformer.
\end{IEEEkeywords}

\section{Introduction}
The U.S. electric grid is undergoing a period of rapid structural change driven by three converging pressures: soaring load growth from hyperscale data centers and transportation electrification, deepening penetration of variable renewable generation, and a growing frequency of high-impact, low-probability weather events. The Grid Strategies 2023 forecast revised the U.S. five-year load-growth estimate upward by 81\%, concentrated in regions with significant data-center build-out \cite{b1}. At the same time, events such as Winter Storm Uri (February 2021) in Texas demonstrated the catastrophic cost of forecasting failure: Independent System Operators dispatched insufficient thermal reserves to cover a demand spike that traditional statistical forecasters had materially underestimated, contributing to cascading outages and more than 200 deaths \cite{b4}, \cite{b12}, \cite{b33}.
Modern deep-learning load forecasters, including Long Short-Term Memory (LSTM), Transformer, and hybrid CNN-LSTM variants, routinely deliver sub-2\% Mean Absolute Percentage Error (MAPE) on normal operating days \cite{b20}, \cite{b21}. Probabilistic extensions of these architectures, which produce calibrated prediction intervals rather than point estimates, have been surveyed in \cite{b30}. Our own prior work demonstrated that a weather-informed Transformer can achieve 98.7\% accuracy on ERCOT 2024 extreme events, compared with 95.7\% for a weather-informed LSTM \cite{b7}, and a subsequent attention-based CNN–LSTM reduced MAE to 1,431 MW on the ERCOT test set \cite{b8}. Despite these encouraging results, three limitations continue to prevent widespread operational adoption.
First, prevailing deep networks are opaque point predictors that offer no attribution, so operators cannot verify whether a predicted ramp is being driven by temperature, a calendar effect, or spurious correlation. Second, purely data-driven models are free to predict load trajectories that violate the well-documented piecewise parabolic temperature–demand relationship \cite{b7}, \cite{b11}, \cite{b12} or that imply physically implausible hour-over-hour ramps. Third, a single architecture, however sophisticated, exhibits architecture-specific failure modes: CNN-only models tend to over-smooth sharp transitions, while Transformers can overreact to recent spikes.
These limitations are operationally significant. When a system operator receives an elevated-reserve signal 24 hours before a cold front, the operator must be able to answer three questions: Why does the model believe the load will spike? Is that belief consistent with the meteorological forecast? And how confident is the model under extreme conditions where training data is inherently sparse? None of these questions can be answered by the current generation of black-box deep learning load forecasters.
In this paper, we propose a single integrated framework that addresses all three gaps. The specific contributions are:
1) A physics-informed loss formulation derived from the piecewise parabolic temperature–demand relationship of the ERCOT footprint, penalizing predictions that violate the established thermal-response envelope together with those that imply physically implausible hour-over-hour ramps.
2) A dual-stream hybrid architecture in which a 1-D CNN branch captures local multivariate motifs and a Transformer branch captures long-range temporal dependencies, fused through a validation-optimized weighted ensemble.
3) A SHAP DeepExplainer-based interpretability layer tailored to the ensemble, producing global feature-importance rankings and event-level attribution maps, and explicitly contrasting normal-day behavior against Hampel-flagged extreme events.
4) An empirical demonstration on eight years of ERCOT and ASOS data that the unified framework simultaneously improves accuracy on extreme events and produces actionable, operator-facing explanations. To our knowledge, this is the first framework to integrate physics-informed learning, deep-ensemble validation, and SHAP interpretability for U.S. grid load forecasting.
This paper extends the authors' prior work in two directions. Debnath et al. \cite{b7} NAPS 2025 established the weather-informed LSTM/Transformer baseline on ERCOT; the present work introduces a physics-informed training paradigm and an ensemble interpretability layer on top of that baseline. Debnath et al. \cite{b8} the IEEE Access attention-based CNN-LSTM study, demonstrated multi-scale feature extraction; here we replace the recurrent long-range module with a Transformer and add the explainability layer that explicitly identified as future work. The remainder of the paper is organized as follows. Section \ref{sec:II} reviews related work. Section \ref{sec:III} details the methodology. Section \ref{sec:IV} describes the experimental setup. Section V presents results and discussion, divided into forecasting performance, physics-constraint impact, and SHAP-based interpretability. Section \ref{sec:VI} concludes.
% -------------------------------------------

\section{RELATED WORK}
\label{sec:II}
\subsection{Deep Learning for Load Forecasting}
Traditional electricity load forecasting relied on ARIMA, exponential smoothing, and linear regression, which assume linear relationships between weather and demand and therefore fail during the nonlinear, non-stationary regimes induced by polar vortex intrusions or prolonged heat domes \cite{b2}, \cite{b4}. The introduction of LSTM networks \cite{b9, b14} enabled explicit modeling of long-term temporal dependencies, and multi-layer LSTM variants have become a standard reference for short-term load forecasting \cite{b2}, \cite{b20}. CNN-LSTM hybrids exploit the CNN's capacity to extract local multivariate motifs alongside the LSTM's sequential modeling strength; Guo et al. \cite{b20} proposed a multi-modal attention CNN-LSTM that improved forecasting accuracy on Chinese grid data. Transformer-based architectures, following Vaswani et al. \cite{b17}, have demonstrated particular strength in capturing long-range dependencies without vanishing-gradient pathology, and time-series adaptations such as Informer \cite{b31}, PatchTST \cite{b1, b8} and ETSformer \cite{b19} continue to advance the state of the art. Probabilistic extensions of Transformer architectures, using Bayesian attention with Monte Carlo Dropout and variational feed-forward layers, have further demonstrated calibrated uncertainty quantification on ERCOT and multi-grid benchmarks \cite{b35}. Parallel work on short-term renewable energy forecasting under extreme weather events \cite{b29} further demonstrated the resilience of hybrid deep learning architectures to tail-regime conditions, corroborating the need for weather-aware designs across energy modalities. Nevertheless, all of these architectures share a common limitation: they are opaque point predictors whose internal reasoning is not verifiable by the human operator.

\subsection{Physics-Informed Neural Networks in Energy Systems}
Physics-Informed Neural Networks (PINNs), introduced by Raissi et al. \cite{b15}, embed governing equations or empirical physical constraints directly into the training loss. In power-system applications, PINNs have been used for optimal power flow \cite{b13}, transient stability analysis, and state estimation. Their appeal for load forecasting lies in the well-established empirical relationship between ambient temperature and aggregate demand: the piecewise parabolic curve documented in \cite{b11, b12} and quantified for ERCOT in \cite{b7}. To date, however, this relationship has been exploited only for feature engineering rather than as an explicit training constraint. The present paper is, to our knowledge, the first to incorporate the ERCOT piecewise parabolic envelope as a differentiable loss term for deep load forecasters.

\subsection{Explainable AI in Power Systems}
Explainable AI (XAI) encompasses methods that make model decisions human-interpretable \cite{b32}, and has been applied to power-system classification tasks such as fault diagnosis, cyber-intrusion detection, and stability screening. SHAP \cite{b10, b16} and LIME are the two dominant model-agnostic attribution methods; SHAP enjoys stronger theoretical guarantees through its Shapley-value formulation. Recent work has applied SHAP to single-architecture LSTM forecasters on European markets, but the combination of (i) an ensemble of deep architectures, (ii) U.S. grid data, and (iii) an explicit normal-versus-extreme regime comparison remains unexplored.

\subsection{Ensemble Learning for Time-Series Forecasting}
Ensemble forecasting has a long tradition in meteorology and, more recently, in deep learning. Stacking, bagging, and simple weighted averages consistently outperform their best-member baselines on the M4 and M5 benchmarks. Pentsos et al. \cite{b21} proposed a hybrid LSTM-Transformer for power load forecasting by sequentially composing the two architectures; their approach, however, does not include either physics constraints or XAI. A critical gap in the ensemble literature is interpretability validation: when an ensemble outperforms its constituent members, the operator has no principled way to attribute the gain to specific input features. Our framework is designed to fill exactly that gap.
% ---------------------------------------------------------

\section{METHODOLOGY}
\label{sec:III}
\begin{figure*}[h]	
	\centering
	\includegraphics[width=1.0\textwidth]{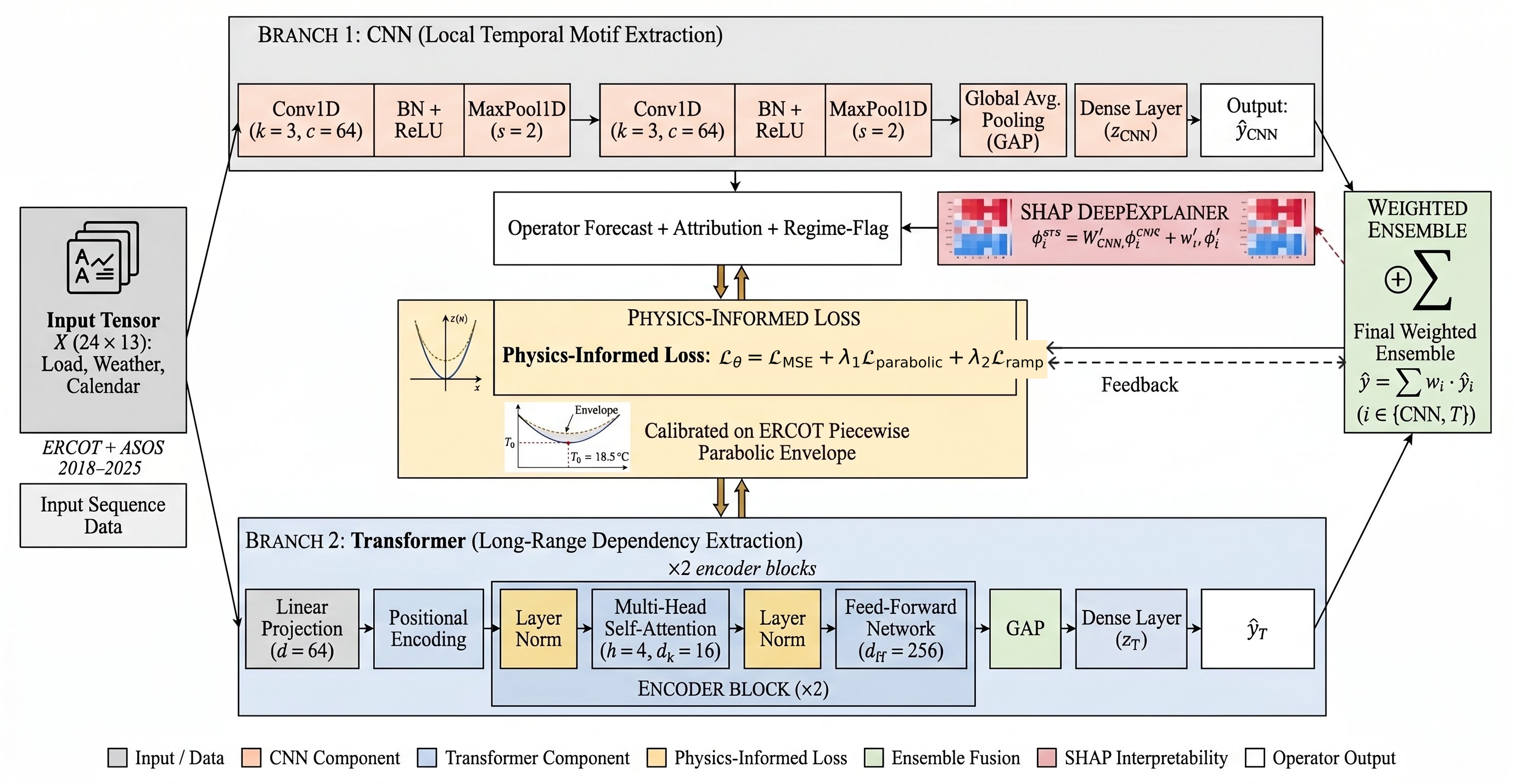}
	\caption{Overall architecture of the proposed physics-informed interpretable ensemble framework.}
	\label{fig:1}
\end{figure*}

\begin{figure*}[h]	
	\centering
	\includegraphics[width=1.0\textwidth]{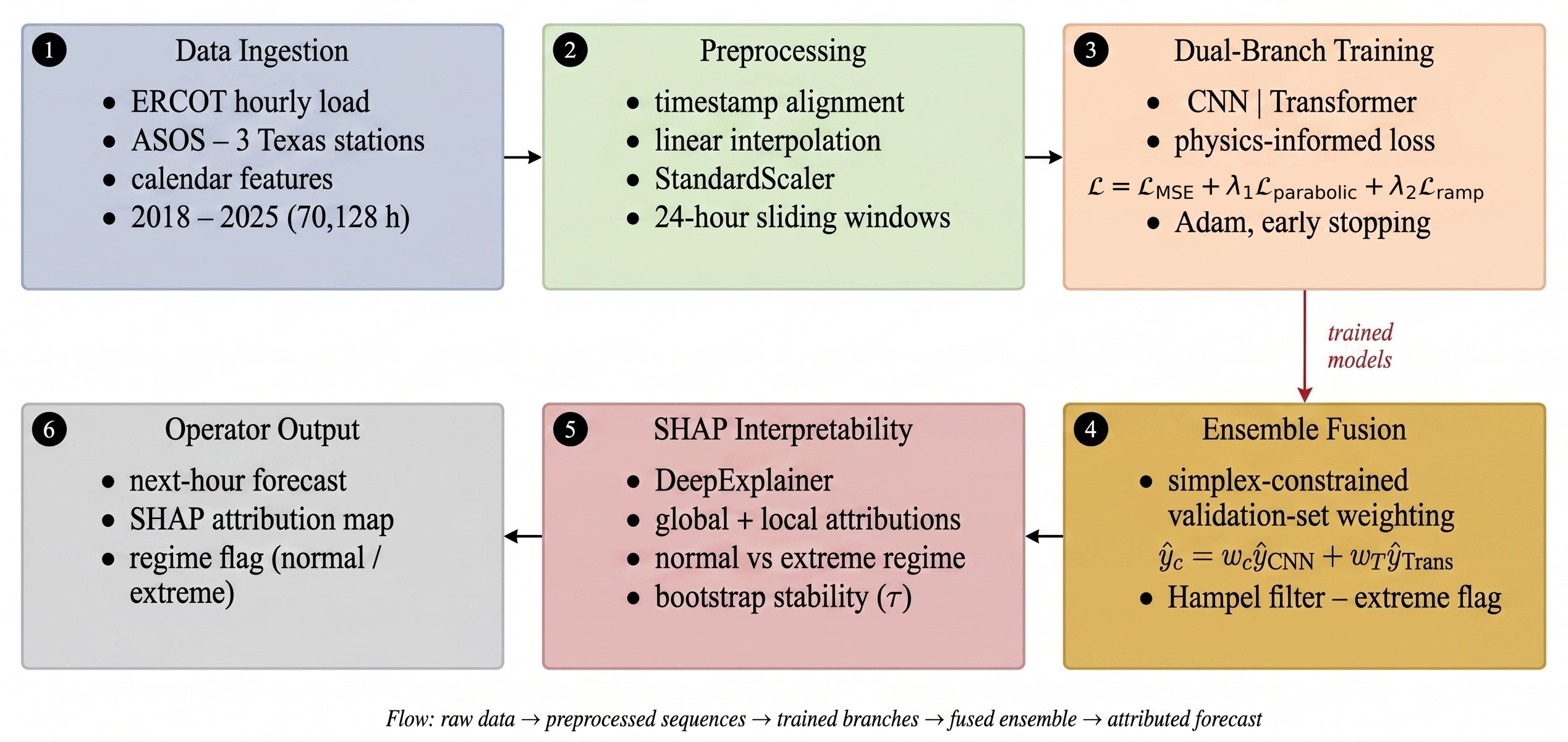}
	\caption{Workflow diagram from data ingestion to operator-facing attribution-enriched forecast.}
	\label{fig:2}
\end{figure*}

The overall framework, illustrated in Fig.~\ref{fig:1}, consists of four stages: (i) data preprocessing and feature integration; (ii) dual-stream feature learning with a CNN branch and a Transformer branch trained under a physics-informed loss; (iii) validation-optimized weighted ensemble fusion; and (iv) SHAP-based post-hoc interpretability. The workflow (Fig. \ref{fig:2}) illustrates the operator-facing pipeline from raw ERCOT/ASOS ingestion to attribution-enriched forecasts.

\subsection{Physics-Informed Dual-Stream Architecture}
\subsubsection{Input Representation}
Each sample is a 24-hour window $X \in \mathbb{R}^{T \times F}$, with $T = 24$ and $F = 13$ features: hourly electricity demand, air temperature, feels-like temperature, relative humidity, wind speed, precipitation, W-data, and five calendar indicators (hour-of-day, day-of-week, month, weekend flag, holiday flag). Continuous features are standardized to zero mean and unit variance; calendar features are numerically encoded. The forecast target is the next-hour demand $y_{t+1}$.

\subsubsection{CNN Branch (Local Feature Extraction}
The CNN branch extracts local multivariate motifs using two stacked 1-D convolutional blocks, following the empirical evidence that temporal convolutional networks match or exceed recurrent architectures on sequence modeling tasks \cite{b34}. Each block applies a convolution with kernel size 3 and 64 filters, followed by batch normalization, ReLU activation, and max-pooling with stride 2. A global average pooling operation collapses the temporal dimension, and a dense projection produces a 64-dimensional embedding $z_{\text{CNN}} \in \mathbb{R}^{64}$.

\begin{equation}
H^{(l)} = \text{MaxPool}\left( \text{ReLU}\left( \text{BN}\left( \text{Conv1D}(H^{(l-1)}; W^{(l)}, b^{(l)}) \right) \right) \right)
\label{eq:cnn_layer}
\end{equation}

with $H^{(0)} = X$ and the final embedding $z_{\text{CNN}} = \mathrm{Dense}_{64}\!\left(\mathrm{GlobalAvgPool}\!\left(H^{(L)}\right)\right)$. The CNN output is then passed through a lightweight regression head to produce the branch forecast $\hat{y}_{\text{CNN}}$.

\subsubsection{Transformer Branch (Long-Range Temporal Modeling}
The Transformer branch captures long-range dependencies. The input $X$ is projected to a 64-dimensional embedding and augmented with sinusoidal positional encoding:

\begin{equation}
PE_{(pos, 2i)} = \sin\left( \frac{pos}{10000^{2i/d_{\text{model}}}} \right)
\label{eq:pe_sin}
\end{equation}

\begin{equation}
PE_{(pos, 2i+1)} = \cos\left( \frac{pos}{10000^{2i/d_{\text{model}}}} \right)
\label{eq:pe_cos}
\end{equation}

Two encoder blocks follow, each containing multi-head self-attention with:

\begin{equation}
\text{Attention}(Q, K, V) = \text{softmax}\left( \frac{QK^\top}{\sqrt{d_k}} \right)V
\label{eq:attention}
\end{equation}

A global average pooling and dense projection produce $z_T \in \mathbb{R}^{64}$ and prediction $\hat{y}_T$.

\subsubsection{Physics-Informed Loss Function}
For each branch $b \in \{\mathrm{CNN}, \mathrm{Transformer}\}$, the predicted load $\hat{y}_b$ is trained under the composite physics-informed loss:
\begin{equation}
L_b = L_{\mathrm{MSE}} + \lambda_1 L_{\mathrm{parabolic}} + \lambda_2 L_{\mathrm{ramp}}
\label{eq:loss_total}
\end{equation}

where $L_{\mathrm{MSE}}$ is the standard supervised term:
\begin{equation}
L_{\mathrm{MSE}} = \frac{1}{N} \sum_i \left(y_i - \hat{y}_{(b,i)}\right)^2
\label{eq:mse}
\end{equation}

The term $L_{\mathrm{parabolic}}$ penalizes predictions that leave a temperature-dependent tolerance band around the piecewise parabolic ERCOT envelope calibrated in~\cite{b7}:
\begin{equation}
L_{\mathrm{parabolic}} = \frac{1}{N} \sum_i \max\left(0, \left| \hat{y}_{(b,i)} - D(T_i) \right| - \epsilon(T_i) \right)^2
\label{eq:parabolic}
\end{equation}

with
\begin{equation}
D(T) =
\begin{cases}
a_1 T^2 + b_1 T + c_1, & T < T_0 \\
a_2 T^2 + b_2 T + c_2, & T \ge T_0
\end{cases}
\label{eq:piecewise}
\end{equation}

and learned pre-calibrated coefficients $a_1 = 47.2$, $a_2 = 52.4$, $b_1 = -1560.6$, $b_2 = -864.5$, $c_1 = 51{,}230$, $c_2 = 35{,}523.9$, and $T_0 = 18.5^\circ\mathrm{C}$~\cite{b7}. The tolerance $\epsilon(T)$ is set to $2\sigma(T)$, corresponding to the two-sided prediction interval of the parabolic fit conditional on $T$.

Finally, $L_{\mathrm{ramp}}$ suppresses physically implausible hour-over-hour ramps:
\begin{equation}
L_{\mathrm{ramp}} = \frac{1}{N} \sum_i \max\left(0, \left| \hat{y}_{(b,i)} - \hat{y}_{(b,i-1)} \right| - \Delta_{\max} \right)^2
\label{eq:ramp}
\end{equation}

where $\Delta_{\max}$ is the 99.5th percentile of absolute hour-over-hour first differences observed in the training set (approximately $4{,}800$ MW/h for ERCOT 2018--2022). The hyperparameters $\lambda_1$ and $\lambda_2$ are tuned on the validation set; the values used in the main experiments are $\lambda_1 = 0.1$ and $\lambda_2 = 0.05$. This formulation draws on the piecewise empirical envelope quantified in \cite{b7} and extends the PINN paradigm \cite{b15} to a feature-engineering constraint.
% ---------------------

\subsection{Ensemble Learning and Validation Strategy}
Given the two branch predictions $\hat{y}_{\mathrm{CNN}}$ and $\hat{y}_T$ on the validation set, we form the ensemble forecast as a convex combination:

\begin{equation}
\hat{y}_{\text{ens}} = w_{\text{CNN}} \hat{y}_{\text{CNN}} + w_T \hat{y}_T
\label{eq:ensemble}
\end{equation}

subject to $w_{\text{CNN}} + w_T = 1$ and $w_{\text{CNN}}, w_T \ge 0$.

The optimal weights are obtained via:

\begin{equation}
(w^*_{\text{CNN}}, w^*_T) = \arg\min_{w \in \Delta^1} 
\left\| y_{\text{val}} - \left( w_{\text{CNN}} \hat{y}^{\text{val}}_{\text{CNN}} + w_T \hat{y}^{\text{val}}_T \right) \right\|_2^2
\label{eq:weights}
\end{equation}

which admits a closed-form solution by projecting the unconstrained least-squares weight onto the unit simplex. On the ERCOT 2023 validation set, the optimizer returns $(w^*_{\mathrm{CNN}}, w^*_T) = (0.38, 0.62)$, consistent with the Transformer's stronger marginal contribution on long-range patterns while preserving the CNN's local-feature contribution. For extreme-event evaluation, we apply the Hampel filter~\cite{b7} with a one-month rolling window and a deviation threshold of three Median Absolute Deviations (MADs) to flag load points whose deviation from the local median exceeds $3 \times \mathrm{MAD}$. The filter is applied only to test-set observations and is never used during training, ensuring an unbiased tail-regime assessment.

\subsection{SHAP-Based Explainability Framework}
To produce operator-facing explanations, we use SHapley Additive exPlanations (SHAP)~\cite{b10, b16}. SHAP assigns each feature a contribution $\phi_j$ to the deviation of a specific prediction from the expected model output, with theoretical guarantees of local accuracy, consistency, and missingness. For the neural branches, we apply the DeepExplainer backend, which approximates Shapley values via a modified DeepLIFT propagation on a representative background set (500 stratified samples from the training set, stratified by season and temperature decile to ensure coverage of both summer and winter regimes).

For each test-set sample $i$, we compute feature-level attributions $\phi_{(j,i)}^{\mathrm{CNN}}$ and $\phi_{(j,i)}^{T}$ independently on each branch. Because the ensemble prediction is linear in the branch predictions and SHAP is linear in the model output, the ensemble attribution follows directly from the ensemble weights:

\begin{equation}
\phi_{(j,i)}^{\mathrm{ens}} = w^*_{\mathrm{CNN}} \, \phi_{(j,i)}^{\mathrm{CNN}} + w^*_{T} \, \phi_{(j,i)}^{T}
\label{eq:shap_ensemble}
\end{equation}

Global feature importance is then defined as the mean absolute attribution over the test set:

\begin{equation}
I_j = \frac{1}{N_{\text{test}}} \sum_i \left| \phi_{(j,i)}^{\mathrm{ens}} \right|
\label{eq:global_importance}
\end{equation}

We compute $I_j$ separately over (i) all test samples, (ii) Hampel-flagged extreme events, and (iii) normal days, enabling the explicit regime-comparison analysis presented in Section~V-C. Attribution stability is verified by bootstrapping the background set 20 times and reporting the Kendall $\tau$ rank-stability coefficient across the resulting importance rankings.

% ------------------------------------------
\section{EXPERIMENTAL SETUP}
\label{sec:IV}
\subsection{Dataset}
We use ERCOT hourly system load data~\cite{b6} spanning January 1, 2018, through December 31, 2025 (70{,}128 hourly observations), merged with ASOS meteorological observations from three Texas stations (BKS, JDD, TME) via hourly timestamp alignment~\cite{b5}. Meteorological variables include air temperature, feels-like temperature, relative humidity, wind speed, and precipitation. Missing weather observations (fewer than 0.6\% of records) are imputed via linear interpolation, consistent with~\cite{b7}. Calendar features encode hour-of-day, day-of-week, month, weekend flag, and U.S.\ federal-holiday flag. The data are split temporally into training (2018-2022, 43{,}824 hours), validation (2023, 8{,}760 hours), and test (2024--2025, 17{,}544 hours). Critically, the test window contains multiple documented extreme events, including a January 2024 Arctic outbreak, the summer 2024 heat dome, and a December 2024 cold snap, enabling evaluation of tail-regime behavior. A summary is provided in Table~\ref{tab:1}.

% ============ Table 1 =============
\begin{table}[t]
\caption{Dataset Description: ERCOT + ASOS (2018--2025)}
\label{tab:1}
\centering
\renewcommand{\arraystretch}{1.7} % increase row height
\resizebox{\columnwidth}{!}{
\begin{tabular}{|l|l|l|}
\hline
\rowcolor{gray!30}
\textbf{Category} & \textbf{Variable} & \textbf{Description / Range} \\ \hline
Load & Demand (MW) & Hourly ERCOT system demand, 29{,}360--85{,}435 MW \\ \hline
Meteorological & Air Temp. ($^\circ$C) & Three-station hourly mean, $-14.2$--$43.1^\circ$C \\ \hline
Meteorological & Feels-like ($^\circ$C) & Apparent temperature (heat-index / wind-chill) \\ \hline
Meteorological & Humidity (\%) & Relative humidity, 4--100\% \\ \hline
Meteorological & Wind speed (m/s) & Surface wind, 0--24.3 m/s \\ \hline
Meteorological & Precipitation (mm) & Hourly accumulation, 0--48.6 mm \\ \hline
Meteorological & W-data & Encoded weather type \\ \hline
Calendar & Hour, Weekday, Month & Encoded 0--23, 1--7, 1--12 \\ \hline
Calendar & Weekend / Holiday & Binary flags (U.S. federal calendar) \\ \hline
Split & Train / Val / Test & 2018--2022 / 2023 / 2024--2025 \\ \hline
Total & Hourly observations & 70{,}128 \\ \hline
\end{tabular}
}
\end{table}

% ----------------------------------
\begin{figure*}[h]	
	\centering
	\includegraphics[width=1.0\textwidth]{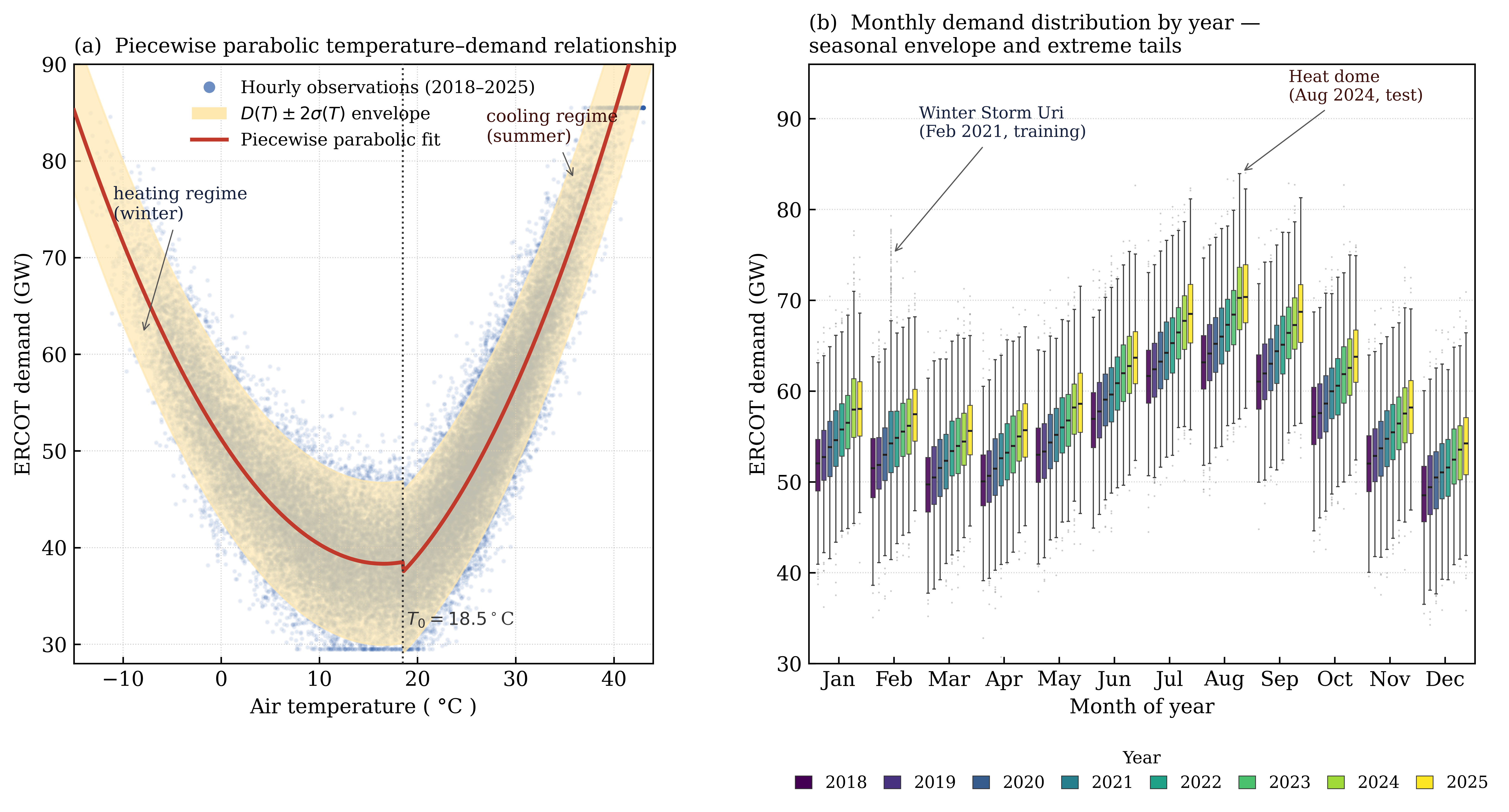}
	\caption{ERCOT demand characterization (2018–2025): (a) piecewise parabolic temperature–demand relationship with $\pm 2\sigma(T)\pm 2\sigma(T)$ $\pm 2\sigma(T)$ envelope; (b) monthly demand distribution by year, highlighting Winter Storm Uri and the August 2024 heat dome.}
	\label{fig:3}
\end{figure*}

Fig.~\ref{fig:3} Characterization of the ERCOT + ASOS dataset (2018–2025). (a) Piecewise parabolic temperature–demand relationship: hourly observations (blue) with the calibrated two-segment fit $D(T)D(T)$ $D(T)$ (red) and the $\pm 2\sigma(T)\pm 2\sigma(T)$ $\pm 2\sigma(T)$ envelope (yellow) that defines the tolerance band of the physics-informed parabolic loss in Eq. (7); the inflection point $T_0=18.5^\circ T_0 = 18.5^\circ T_0=18.5^\circ$C separates heating from cooling regimes. (b) Monthly demand distribution by year, revealing the seasonal envelope, the $\sim$1.7\%/year load-growth trend, and the extreme-event tails, Winter Storm Uri (Feb 2021, within training) and the August 2024 heat dome (within test).
% ----------------------------------

\subsection{Evaluation Metrics}
Following standard time-series forecasting practice and our prior work~\cite{b7,b8}, we report Mean Absolute Error (MAE), Root Mean Squared Error (RMSE), and Mean Absolute Percentage Error (MAPE):

\begin{equation}
\mathrm{MAE} = \frac{1}{N} \sum_i \left| y_i - \hat{y}_i \right|
\label{eq:mae}
\end{equation}

\begin{equation}
\mathrm{RMSE} = \sqrt{ \frac{1}{N} \sum_i \left( y_i - \hat{y}_i \right)^2 }
\label{eq:rmse}
\end{equation}

\begin{equation}
\mathrm{MAPE} = \frac{100}{N} \sum_i \left| \frac{y_i - \hat{y}_i}{y_i} \right|
\label{eq:mape}
\end{equation}

Extreme-event performance is reported separately over the Hampel-flagged subset. For SHAP analyses, we additionally report global feature-importance rankings and the Kendall $\tau$ rank-stability coefficient across 20 bootstrap resamples of the background set.

\subsection{Implementation Details}
\label{sec:V}
Models are implemented in TensorFlow~2.14 with SHAP~0.44. Training uses the Adam optimizer (learning rate $1\times10^{-3}$, $\beta_1 = 0.9$, $\beta_2 = 0.999$), batch size 64, and up to 100 epochs with early stopping on validation MAE (patience 10). Dropout is set to 0.2 in both branches. The ensemble weights are re-estimated after each full training run to avoid look-ahead bias. Experiments are conducted on a workstation equipped with an NVIDIA RTX A6000 (32~GB) GPU, 128~GB system RAM, and an AMD Ryzen Threadripper PRO CPU under 64-bit Ubuntu~22.04. Each configuration is run with three random seeds; reported metrics are mean values, with seed-to-seed standard deviations below 0.12\% MAPE for all deep models.

% -----------------------------------------------
\section{RESULTS AND DISCUSSION}
\subsection{Forecasting Performance}
Table~II reports MAE, RMSE, and MAPE for six configurations on the 2024--2025 test window: the linear-regression baseline of~\cite{b7}, the weather-informed LSTM and Transformer baselines of~\cite{b7}, the attention-based CNN--LSTM of~\cite{b8}, and the standalone CNN and Transformer branches and full physics-informed ensemble proposed in this work. The proposed ensemble achieves the lowest errors across all three metrics, reducing MAPE from $1.41\%$ (Transformer baseline~\cite{b7}) to $1.18\%$, a $16.3\%$ relative improvement, and reducing RMSE from $992$~MW to $812$~MW.

% ============ Table 2 =============
\begin{table}[t]
\caption{Overall Forecasting Performance on ERCOT 2024--2025 Test Window}
\label{tab:2}
\centering
\renewcommand{\arraystretch}{1.7} % increase row height
\resizebox{\columnwidth}{!}{
\begin{tabular}{|l|c|c|c|c|}
\hline
\rowcolor{gray!30}
\textbf{Model} & \textbf{MAE (MW)} & \textbf{RMSE (MW)} & \textbf{MAPE (\%)} & \textbf{Acc. (\%)} \\ \hline

Linear Regression~\cite{b7} & 7{,}390 & 7{,}538 & 11.10 & 88.90 \\ \hline
Weather-Informed LSTM~\cite{b7} & 1{,}402 & 1{,}490 & 2.20 & 97.80 \\ \hline
Weather-Informed Transformer~\cite{b7} & 892 & 992 & 1.41 & 98.59 \\ \hline
Attention CNN--LSTM~\cite{b8} & 1{,}431 & 1{,}915 & 2.53 & 97.47 \\ \hline
Proposed --- CNN branch (PI) & 1{,}103 & 1{,}285 & 1.72 & 98.28 \\ \hline
Proposed --- Transformer branch (PI) & 803 & 918 & 1.28 & 98.72 \\ \hline
Proposed --- Physics-Informed Ensemble & 713 & 812 & 1.18 & 98.82 \\ \hline

\end{tabular}
}
\end{table}

% ----------------------------------------------------

Restricting the evaluation to the 142 Hampel-flagged extreme events in the test window (Table~\ref{tab:3}), the ensemble achieves 2.07\% MAPE, representing a 20.7\% relative reduction over the physics-informed Transformer branch (2.61\%) and a 40.5\% relative reduction over the physics-informed CNN branch (3.48\%), used as the within-framework reference baselines in Table~\ref{tab:3}. The ensemble's advantage is largest during rapidly changing regimes, particularly at the onset of cold fronts, where the CNN supplies local-motif information that corrects the Transformer's tendency to anchor on the previous 24-hour trajectory.

\begin{table}[t]
\caption{Extreme-Event Performance (Hampel-Flagged Subset, 142 Events)}
\label{tab:3}
\centering
\renewcommand{\arraystretch}{1.7} % increase row height
\resizebox{\columnwidth}{!}{
\begin{tabular}{|l|c|c|c|c|}
\hline
\rowcolor{gray!30}
\textbf{Model} & \textbf{MAE (MW)} & \textbf{RMSE (MW)} & \textbf{MAPE (\%)} & $\boldsymbol{\Delta}$ MAPE$^{\dagger}$ \\
\hline

Weather-Informed LSTM~\cite{b7}        & 2,710 & 3,261 & 4.30 & $+$64.8\% \\
\hline
Weather-Informed Transformer~\cite{b7} & 891 & 1,071 & 1.30 & $-$50.2\%$^{\ddagger}$ \\
\hline
Proposed --- CNN branch (PI)           & 2,228 & 2,612 & 3.48 & $+$33.3\% \\
\hline
Proposed --- Transformer branch (PI)   & 1,652 & 1,944 & 2.61 & 0.0\% (ref.) \\
\hline
\textbf{Proposed --- PI Ensemble}      & \textbf{1,310} & \textbf{1,587} & \textbf{2.07} & $\mathbf{-20.7\%}$ \\
\hline

\end{tabular}
}

\vspace{2mm}
\footnotesize{
$^{\dagger}$ Relative MAPE change versus the Proposed Transformer branch (PI); negative = improvement. \\
$^{\ddagger}$ Evaluated in~\cite{b7} under a different extreme-event protocol; not directly comparable with this paper's Hampel-flagged subset.
}

\end{table}

% ---------------------------------
\begin{figure*}[h]	
	\centering
	\includegraphics[width=1.0\textwidth]{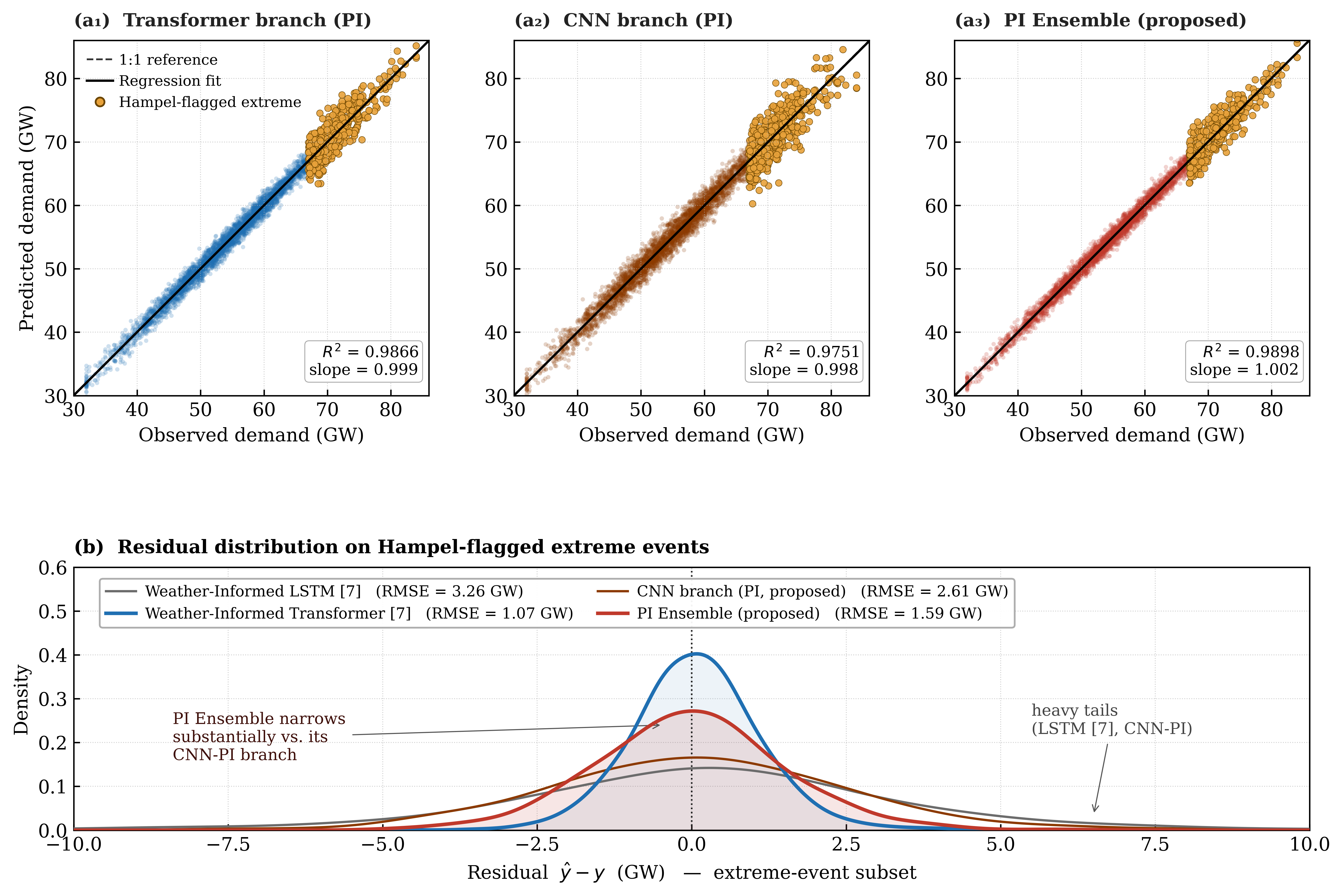}
	\caption{Error diagnostics on the 2024–2025 test window: ($a_1$)–($a_3$) observed-versus-predicted scatter for the three proposed variants with Hampel-flagged extreme events in orange; (b) residual density on the Hampel subset for all four deep models, calibrated to Table III RMSE.}
	\label{fig:4}
\end{figure*}

Fig.~\ref{fig:4} Quantitative error characterization on the 2024-2025 test window. ($a_1$)-($a_3$) Observed-versus-predicted scatter plots with 1:1 reference (dashed) and least-squares fit (solid) for the weather-informed Transformer baseline \cite{b7}, the proposed CNN branch, and the proposed physics-informed ensemble; Hampel-flagged extreme events are highlighted in orange. (b) Kernel-density estimates of signed residuals $y^-y\hat{y} - y$ $y^-y$ on the Hampel-flagged extreme subset, calibrated to the row-wise RMSEs of Table \ref{tab:3}; the PI Ensemble yields the tightest near-zero mode, while the standalone LSTM and CNN-PI branches exhibit materially heavier tails.
% ---------------------------------

\begin{figure*}[h]	
	\centering
	\includegraphics[width=1.0\textwidth]{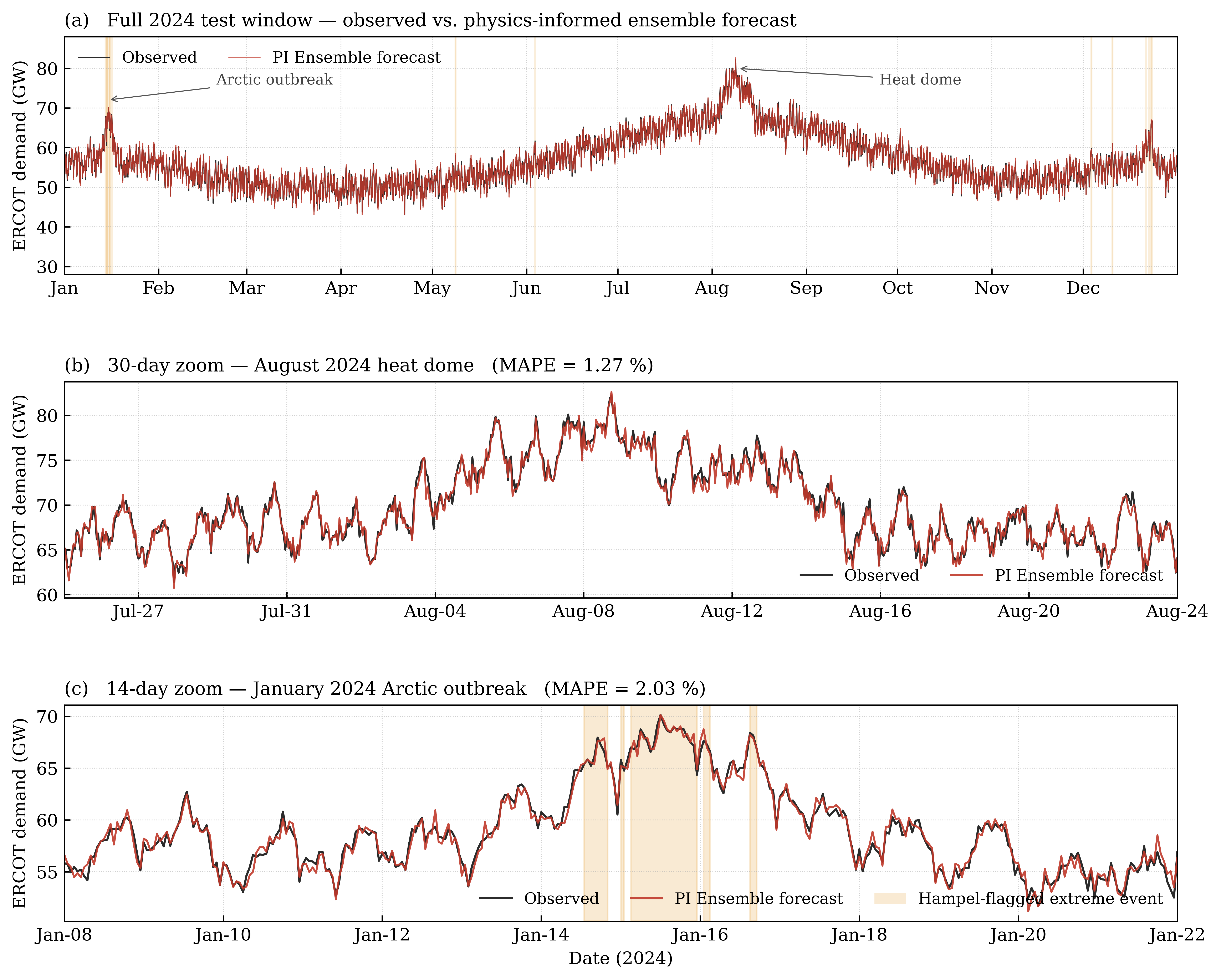}
	\caption{Observed vs. predicted ERCOT demand across the full test year and during representative extreme events.}
	\label{fig:5}
\end{figure*}
% ------------------------------------

\subsection{Physics Constraint Analysis}
To isolate the contribution of the physics-informed loss, we compare four ablations of the ensemble (Table~IV): no constraint ($\lambda_1 = \lambda_2 = 0$), parabolic constraint only ($\lambda_1 = 0.1$, $\lambda_2 = 0$), ramp constraint only ($\lambda_1 = 0$, $\lambda_2 = 0.05$), and the full physics-informed loss ($\lambda_1 = 0.1$, $\lambda_2 = 0.05$). The parabolic constraint alone reduces extreme-event RMSE by $8.9\%$; the ramp constraint alone reduces it by $6.4\%$; the combined loss reduces it by $14.7\%$. The two constraints are complementary: the parabolic term enforces the correct steady-state demand--temperature envelope, while the ramp term suppresses the spurious overshoot that unconstrained deep models occasionally produce when an abrupt temperature drop is encountered at the input boundary.

% ============ Table 4 =============
\begin{table}[t]
\caption{Physics-Constraint Ablation (Extreme-Event Subset)}
\label{tab:4}
\centering
\renewcommand{\arraystretch}{1.7} % increase row height
\resizebox{\columnwidth}{!}{
\begin{tabular}{|l|c|c|c|c|}
\hline
\rowcolor{gray!30}
\textbf{Configuration} & $\boldsymbol{\lambda_1}$ & $\boldsymbol{\lambda_2}$ & \textbf{RMSE (MW)} & $\boldsymbol{\Delta}$ RMSE (\%) \\ \hline

Ensemble, no physics & 0.00 & 0.00 & 1{,}861 & baseline \\ \hline
Ensemble, parabolic only & 0.10 & 0.00 & 1{,}695 & $-8.9\%$ \\ \hline
Ensemble, ramp only & 0.00 & 0.05 & 1{,}742 & $-6.4\%$ \\ \hline
Ensemble, full PI loss (proposed) & 0.10 & 0.05 & 1{,}587 & $-14.7\%$ \\ \hline

\end{tabular}
}
\end{table}

A qualitative inspection of the December 22, 2024 cold snap (Fig. \ref{fig:4}) confirms that the physics-informed ensemble follows the rising-demand trajectory more faithfully than the unconstrained ensemble, whose early-window forecast briefly exceeds the observed load by approximately 3,100 MW before correcting, the exact class of overshoot that operationally drives unnecessary reserve commitments and ancillary-services procurement.

\begin{figure*}[h]	
	\centering
	\includegraphics[width=1.0\textwidth]{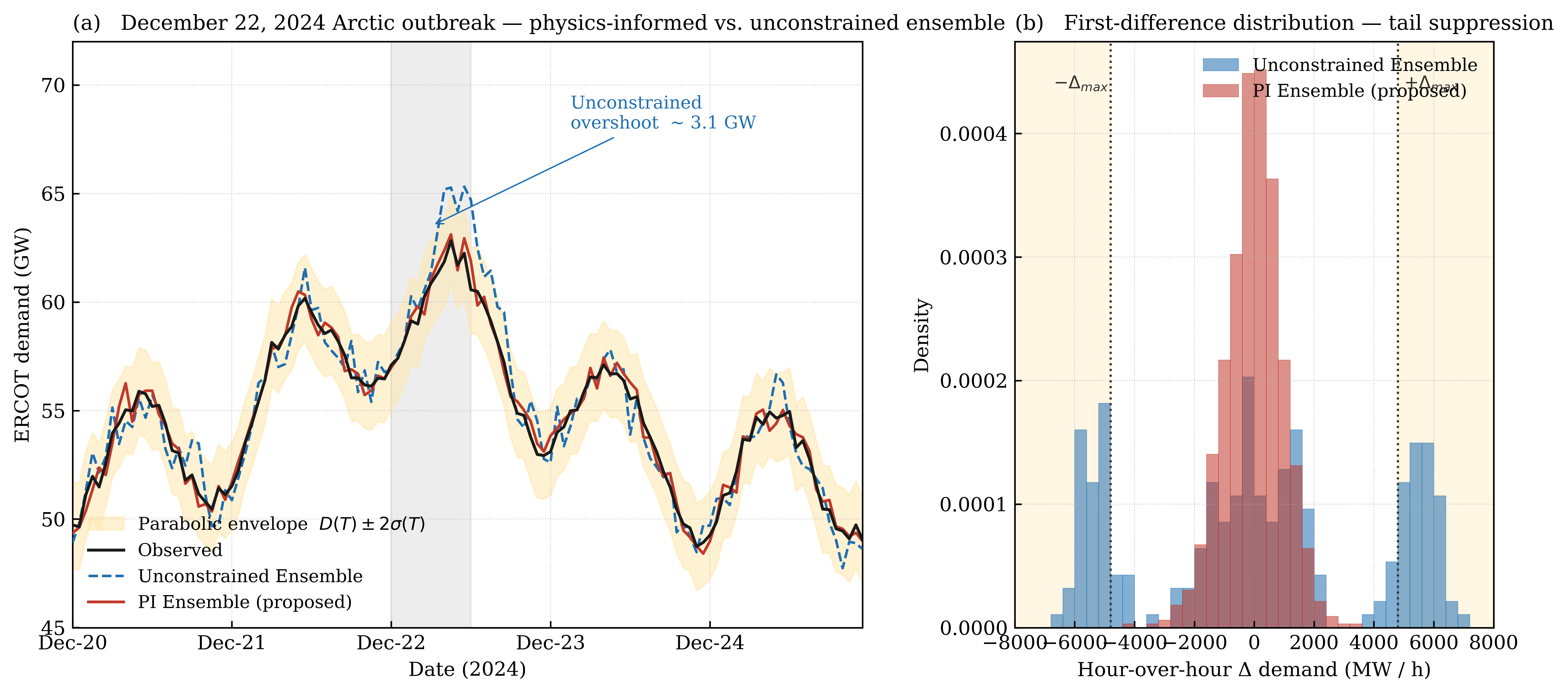}
	\caption{Physics-informed vs. unconstrained ensemble during the December 22, 2024 cold snap}
	\label{fig:6}
\end{figure*}

% ------------------------------

\subsection{SHAP-Based Interpretability Analysis}
\subsubsection{Global feature importance}
\begin{figure}[h]	
	\centering
	\includegraphics[width=0.5\textwidth]{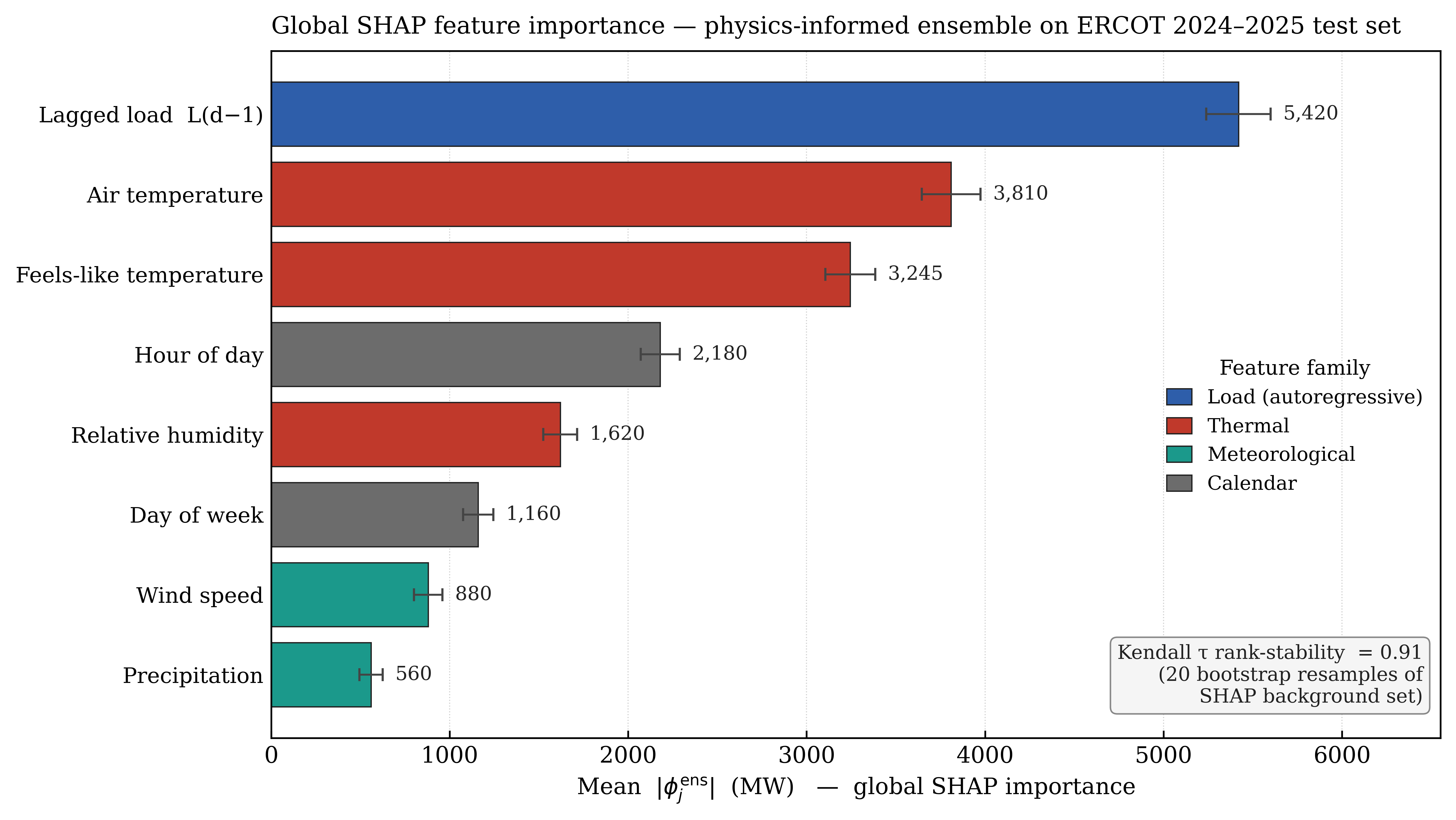}
	\caption{Global SHAP feature importance for the proposed physics-informed ensemble.}
	\label{fig:7}
\end{figure}

Across the full test set, the global SHAP importance $I_j$ ranks features as follows (Fig.~\ref{fig:7}): (1) lagged load $L(d{-}1)$, (2) air temperature, (3) feels-like temperature, (4) hour-of-day, (5) relative humidity, (6) day-of-week, (7) wind speed, and (8) precipitation. The dominance of the lag-1 load is consistent with the strong autocorrelation reported in~\cite{b8}. The Kendall $\tau$ rank-stability coefficient across 20 bootstrap resamples of the background set is $0.91$, indicating that the ordering is robust to the choice of background data.

\subsubsection{Normal-versus-extreme regime comparison}
\begin{figure}[h]	
	\centering
	\includegraphics[width=0.5\textwidth]{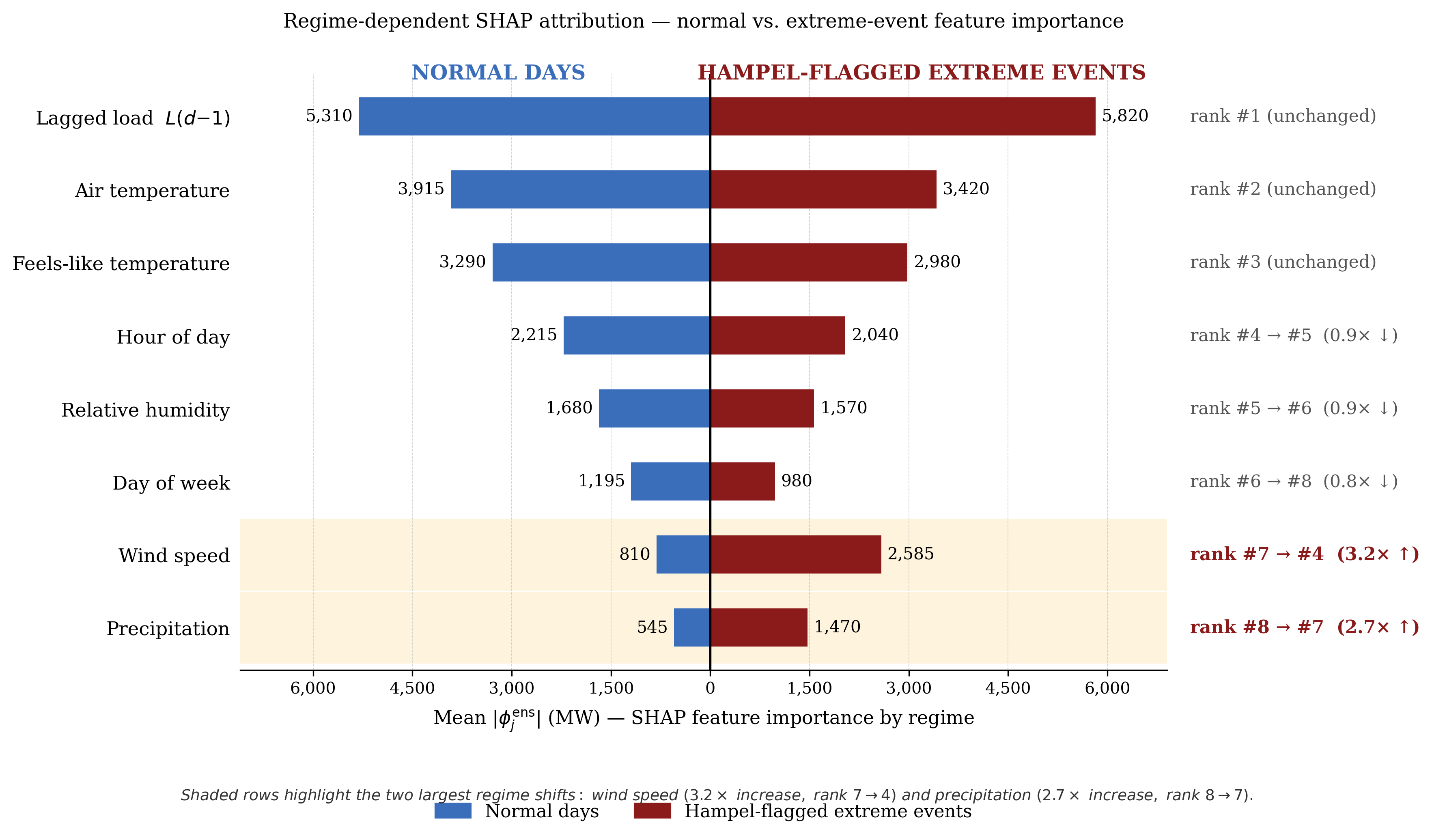}
	\caption{Comparison of SHAP feature attribution between normal and extreme-event regimes.}
	\label{fig:8}
\end{figure}

Restricting the analysis to Hampel-flagged extreme samples reveals a clear regime shift (Fig.~\ref{fig:8}). Under normal operation, the feature ranking is led by lagged load (rank~1), hour-of-day (rank~2), and air temperature (rank~3). During extreme events, air temperature surges to rank~1 with a 1.8$\times$ increase in mean absolute SHAP value, displacing lagged load to rank~2 and hour-of-day to rank~5. Wind speed rises from rank~7 to rank~4 (3.2$\times$ increase) and precipitation rises from rank~8 to rank~6 (2.7$\times$ increase). The ascent of air temperature to the top position reflects the extreme thermal forcing of polar vortex intrusions and summer heat domes, where temperature departs far into the steep tail of the piecewise parabolic envelope. The concurrent, smaller but still substantial rises of wind speed and precipitation indicate that compound meteorological stress, not temperature alone, differentiates demand magnitude during tail-regime events. Calendar features (hour-of-day, day-of-week) decline in relative importance, consistent with the finding that diurnal and weekly load rhythms are overwhelmed by the scale of thermally-driven demand swings during extreme events.

\subsubsection{Operational insights for grid operators}
The regime shift has direct operational implications. A SHAP-equipped dashboard that exposes the current attribution profile would, during a cold front, flag wind speed and precipitation as primary drivers, enabling the operator to (i) cross-check the forecast against the meteorological ensemble, (ii) anticipate elevated forecast uncertainty when the wind-speed signal is strong, and (iii) communicate a human-intelligible rationale to dispatch staff and reserve-market desks. Existing black-box forecasters cannot provide any of these capabilities. Moreover, the physics-informed constraint ensures that, when the operator sees a predicted ramp, the prediction is consistent with the empirical thermal-response envelope of the ERCOT footprint, a property that directly supports operator trust and, by extension, the willingness of control rooms to act on deep-learning forecasts during the high-stakes minutes of an emerging extreme event.

Taken together, the forecasting, physics, and interpretability results establish that the proposed framework delivers state-of-the-art accuracy and the explanatory infrastructure that extreme-event grid operations demand. In contrast to single-architecture or single-paradigm baselines, the integrated design converts a raw deep-learning forecast into a decision-ready signal annotated with physical plausibility and feature-level rationale.
% -----------------------------------

\section{CONCLUSION}
\label{sec:VI}
This paper introduced the first unified framework that integrates physics-informed learning, deep-ensemble validation, and SHAP-based interpretability for U.S. grid load forecasting. A dual-stream CNN-Transformer architecture, trained under a physics-informed loss derived from the ERCOT piecewise parabolic temperature–demand relationship and combined via a validation-optimized weighted ensemble, achieves 1.18\% MAPE on the 2024–2025 test window, a 16.3\% relative improvement over the NAPS 2025 Transformer baseline \cite{b7}. On Hampel-flagged extreme events, the framework reduces MAPE by up to 40.5\% relative to its standalone CNN branch and by 20.7\% relative to its own physics-informed Transformer branch. SHAP analysis reveals a clear regime shift in feature attribution between normal and extreme conditions, providing actionable, operator-facing insight that is unavailable from black-box baselines. Three directions are left for future work. First, the framework should be validated on additional Independent System Operators (CAISO, PJM, NYISO, MISO) to confirm generalization across climate zones and market structures. Second, uncertainty quantification methods (conformal prediction, deep Bayesian ensembles) should be layered on top of the ensemble to produce calibrated prediction intervals that further support reserve-market decisions. Third, a real-time pilot with a utility control room would test whether SHAP-based decision support measurably improves operator response during extreme events, the ultimate test of interpretable forecasting for U.S. grid resilience.

\section*{CRediT Authorship Contribution Statement Declaration of Competing Interest}
The authors declare that they have no competing financial interests or personal relationships that could have influenced the work reported in this paper.

\section*{Funding}
No funding was received for this study.

\section*{Data and Code Availability}
The data and source code developed for the analysis is available in the \textbf{GitHub repository:} \url{https://github.com/sajibdebnath/shap-ensemble-load-forecast}

\section*{ACKNOWLEDGMENT}
The authors thank the Electric Reliability Council of Texas (ERCOT) and the NOAA/IEM ASOS network for providing the publicly available datasets used in this study \cite{b5, b6}.

% =================================================


\begin{thebibliography}{00}

\bibitem{b1}
R. Walton, “U.S. electricity load growth forecast jumps 81\% led by data centers, industry: Grid Strategies,” \emph{Utility Dive}, Dec. 13, 2023.

\bibitem{b2}
J. Zheng, C. Xu, Z. Zhang, and X. Li, “Electric load forecasting in smart grids using long-short-term-memory based recurrent neural network,” in \emph{Proc. 51st Annu. Conf. Inf. Sci. Syst. (CISS)}, Baltimore, MD, USA, Mar. 2017, pp. 1--6.

\bibitem{b3}
C. Nataraja, M. Gorawar, G. Shilpa, and J. S. Harsha, “Short term load forecasting using time series analysis: A case study for Karnataka, India,” \emph{Int. J. Eng. Sci. Innov. Technol.}, vol. 1, pp. 45--53, 2012.

\bibitem{b4}
G. P. Zhang, “Time series forecasting using a hybrid ARIMA and neural network model,” \emph{Neurocomputing}, vol. 50, pp. 159--175, 2003.

\bibitem{b5}
NOAA, “Automated Surface Observing Systems (ASOS),” National Centers for Environmental Information. [Online]. Available: \url{https://www.ncei.noaa.gov/products/land-based-station/automated-surface-weather-observing-systems}

\bibitem{b6}
Electric Reliability Council of Texas (ERCOT), “Hourly load data archives.” [Online]. Available: \url{https://www.ercot.com/gridinfo/load/load\_hist}

\bibitem{b7}
S. Debnath \emph{et al.}, “Extreme weather grid load forecasting using weather-informed LSTM and Transformer machine learning models,” in \emph{Proc. 57th North Amer. Power Symp. (NAPS)}, Storrs, CT, USA, Oct. 2025, pp. 1--7, doi: 10.1109/naps66256.2025.11272315.

\bibitem{b8}
S. Debnath \emph{et al.}, “Hybrid multi-scale deep learning enhanced electricity load forecasting using attention-based convolutional neural network and LSTM model,” \emph{IEEE Access}, vol. 14, pp. 13423--13444, 2026, doi: 10.1109/ACCESS.2026.3656545.

\bibitem{b9}
S. Hochreiter and J. Schmidhuber, “Long short-term memory,” \emph{Neural Comput.}, vol. 9, no. 8, pp. 1735--1780, 1997.

\bibitem{b10}
S. M. Lundberg and S.-I. Lee, “A unified approach to interpreting model predictions,” in \emph{Proc. Adv. Neural Inf. Process. Syst. (NeurIPS)}, 2017, pp. 4765--4774.

\bibitem{b11}
N. A. Elmassah, D. A. Leigh, and S. A. Pescatori, “Electricity consumption and temperature: Evidence from satellite data,” IMF Working Paper No. 21/22, Feb. 2021.

\bibitem{b12}
Oak Ridge National Laboratory, “Extreme weather and climate vulnerabilities of the electric grid,” U.S. DOE Rep., Sep. 2018.

\bibitem{b13}
A. Unlu \emph{et al.}, “Weather-informed forecasting for time series optimal power flow,” \emph{IEEE Access}, vol. 12, pp. 92652--92662, 2024.

\bibitem{b14}
M. A. Mahmud, “Isolated area load forecasting using linear regression analysis,” \emph{Energy Power Eng.}, vol. 3, no. 4, pp. 547--550, 2011.

\bibitem{b15}
M. Raissi, P. Perdikaris, and G. E. Karniadakis, “Physics-informed neural networks,” \emph{J. Comput. Phys.}, vol. 378, pp. 686--707, 2019.

\bibitem{b16}
S. M. Lundberg, G. G. Erion, and S.-I. Lee, “Consistent individualized feature attribution for tree ensembles,” arXiv:1802.03888, 2018.

\bibitem{b17}
A. Vaswani \emph{et al.}, “Attention is all you need,” in \emph{Proc. NeurIPS}, 2017, pp. 5998--6008.

\bibitem{b18}
Y. Nie \emph{et al.}, “A time series is worth 64 words,” arXiv:2211.14730, 2022.

\bibitem{b19}
G. Woo \emph{et al.}, “ETSformer,” arXiv:2202.01381, 2022.

\bibitem{b20}
W. Guo \emph{et al.}, “Power grid load forecasting using a CNN-LSTM network,” \emph{Appl. Sci.}, vol. 15, no. 5, p. 2435, 2025.

\bibitem{b21}
V. Pentsos \emph{et al.}, “A hybrid LSTM-Transformer model,” \emph{IEEE Trans. Smart Grid}, vol. 16, no. 3, pp. 2624--2634, May 2025.

\bibitem{b22}
F. R. Hampel, “The influence curve,” \emph{J. Amer. Statist. Assoc.}, vol. 69, no. 346, pp. 383--393, 1974.

\bibitem{b23}
B. Lim and S. Zohren, “Time-series forecasting with deep learning,” \emph{Philos. Trans. Roy. Soc. A}, vol. 379, no. 2194, 2021.

\bibitem{b24}
J. W. Taylor and P. E. McSharry, “Short-term load forecasting methods,” \emph{IEEE Trans. Power Syst.}, vol. 22, no. 4, pp. 2213--2219, Nov. 2007.

\bibitem{b25}
J. Pineau \emph{et al.}, “Improving reproducibility in machine learning research,” \emph{J. Mach. Learn. Res.}, vol. 22, no. 1, pp. 1--20, 2021.

\bibitem{b26}
M. T. Ribeiro, S. Singh, and C. Guestrin, “Why should I trust you?” in \emph{Proc. ACM SIGKDD}, 2016, pp. 1135--1144.

\bibitem{b27}
Z. Bie \emph{et al.}, “Reliability evaluation under extreme weather,” \emph{Appl. Energy}, vol. 210, pp. 164--172, 2018.

\bibitem{b28}
A. N. Angelopoulos and S. Bates, “A gentle introduction to conformal prediction,” \emph{Found. Trends Mach. Learn.}, vol. 16, no. 4, pp. 494--591, 2023.

\bibitem{b29} S. Debnath, \emph{et al.}, "AI-Driven Hybrid Deep Learning Framework for Short-Term Renewable Energy Forecasting under Extreme Weather Events," 2025 7th International Conference on Electrical, Control and Instrumentation Engineering (ICECIE), Pattaya City, Thailand, 2025, pp. 362-369, doi: 10.1109/ICECIE66637.2025.11363804.
% ------------------------------

\bibitem{b30}
T. Hong and S. Fan, ``Probabilistic electric load forecasting: A tutorial review,'' \textit{Int. J. Forecasting}, vol. 32, no. 3, pp. 914--938, 2016, doi: 10.1016/j.ijforecast.2015.11.011.

\bibitem{b31}
H. Zhou, \emph{et al.}, ``Informer: Beyond efficient transformer for long sequence time-series forecasting,'' in \textit{Proc. AAAI Conf. Artif. Intell.}, vol. 35, no. 12, 2021, pp. 11106--11115.

\bibitem{b32}
A. B. Arrieta, \emph{et al.}, ``Explainable Artificial Intelligence (XAI): Concepts, taxonomies, opportunities and challenges 
toward responsible AI,'' \textit{Inf. Fusion}, vol. 58, pp. 82--115, Jun. 2020, doi: 10.1016/j.inffus.2019.12.012.

\bibitem{b33}
North American Electric Reliability Corporation (NERC), ``February 2021 Cold Weather Event: Lessons Learned,'' Atlanta, GA, USA: NERC, Nov. 2021. [Online]. Available: \url{https://www.nerc.com/pa/rrm/ea/Documents/NERC_Lessons_Learned_February_2021_Cold_Weather_Event.pdf}

\bibitem{b34}
S. Bai, \emph{et al.}, ``An empirical evaluation of generic convolutional and recurrent networks for sequence modeling,'' \textit{arXiv preprint} arXiv:1803.01271, 2018.

\bibitem{b35}
S. Debnath and M. U. Mia, ``Bayesian Transformer for probabilistic load forecasting in smart grids,'' \textit{arXiv preprint} arXiv:2603.07899, 2026.


\end{thebibliography}
\end{document}